\documentclass{article}

\usepackage{spconf}

\usepackage{enumerate}
\usepackage{amsmath, amssymb, amsfonts}
\usepackage{graphicx}
\graphicspath{ {figures/}}
\usepackage{float}
\usepackage{color}

\title{Extendable Neural Matrix Completion}
			
 \name{Duc Minh Nguyen, Evaggelia Tsiligianni, Nikos Deligiannis
                      }
                 \address{Vrije Universiteit Brussel, Pleinlaan 2, B-1050 Brussels, Belgium \\
                          imec, Kapeldreef 75, B-3001 Leuven, Belgium \\
                          \{mdnguyen, etsiligi, ndeligia\}@etrovub.be}

\begin{document}
\ninept
\maketitle	
\begin{abstract}
Matrix completion is one of the key problems in signal processing and machine learning, 
with applications ranging from image processing and data gathering 
to classification and recommender systems.  
Recently, deep neural networks have been proposed as latent factor models 
for matrix completion and have achieved state-of-the-art performance. 
Nevertheless, a major problem with existing neural-network-based models is 
their limited capabilities to extend to samples unavailable at the training stage. 
In this paper, we propose a deep two-branch neural network model for matrix completion. 
The proposed model not only inherits the predictive power of neural networks, 
but is also capable of extending to partially observed samples outside the training set, 
without the need of retraining or fine-tuning. 
Experimental studies on popular movie rating datasets prove the effectiveness of our model 
compared to the state of the art, in terms of both accuracy and extendability.
\end{abstract}

%
\section{Introduction}
\label{sec:intro}
Recovering a matrix from partial observations is a problem of high interest in many 
signal processing and machine learning applications,
where a matrix cannot be fully sampled or directly observed.
Examples of signal processing tasks 
that employ matrix completion algorithms 
include image super resolution~\cite{Cao2014},
image and video denoising ~\cite{Ji2010},
data gathering in wireless sensor networks~\cite{Cheng2013},
and more.
In machine learning, matrix completion has been employed to tackle problems
such as clustering~\cite{yi12}, classification~\cite{luo15},
and recommender systems~\cite{zheng16, monti17, dziugaite15, he17}.

Let  $M \in \mathbb{R}^{n \times m}$ be a matrix with a limited number of observed entries $M_{ij}$,
$(i,j) \in\Omega$,  with  $\Omega$ the set of indices corresponding to the observed entries.
Then, recovering matrix $M$ from the knowledge of the value of its entries in the set  $\Omega$
is formulated as an optimization problem of the form:
\begin{equation}
\label{eq:mc}
R = \arg\min_ {\bar{R}}\|P_{\Omega}(\bar{R}-M)\|_F,
\end{equation}
with $R \in \mathbb{R}^{n \times m}$ denoting the complete matrix,
$P_{\Omega}$ an operator that indexes the entries defined in $\Omega$,
and $\|\cdot \|_F$ the Frobenius norm.

Several studies have focused on the problem of recovering  $R$ from $M$.
Under the assumption that $R$ is a low-rank matrix,
a convex optimization method that solves a nuclear norm minimization problem has been proposed in \cite{candes09}. 
The major drawback of algorithms belonging to this category is their high computational cost, 
especially when the dimensions of the matrix increase.
Low-rank factorization~\cite{Markovsky2012} has been proposed to address large-scale matrix completion problems.
The unknown rank-$r$ matrix is expressed as the product of two much smaller matrices $UV^T$ , 
with $U \in \mathbb{R}^{n \times r}$, $V  \in \mathbb{R}^{m \times r}$, and $r\ll \min(n,m)$, 
so that the low-rank requirement is automatically fulfilled.

Several matrix completion algorithms have been proposed to address the problem of collaborative filtering 
for recommender systems \cite{Koren2009, zheng16, monti17, dziugaite15, he17}.
In this application scenario, the matrix entries reflect users' preferences (ratings) for items. 
Considering low-rank factorization as a mapping of both users and items 
to a joint latent factor space of dimensionality $r$, 
the relation between users and items is modelled as an inner product in that space.
The class of techniques that learn latent representations of users and items 
such that user-item interactions can be modelled as inner products in the latent space 
is referred to as matrix factorization~\cite{Koren2009}.

Recently neural network models have  achieved state-of-the-art performance~\cite{zheng16, monti17, dziugaite15, he17}
in the problem of matrix completion. 
However, a major drawback of existing methods 
is that they cannot be extended to users unseen during training. 
Updating the model at the arrival of new users or items 
as in online matrix completion methods~\cite{jin16, chouvardas17}
can be time consuming, and, thus, impractical in high-dimensional settings.
Moreover, the employment of external information (implicit feedback)
to predict interactions between new users and items---as in~\cite{he17,hsieh17}---may not be applicable in many use cases.
Considering that recommender systems often have to process matrices of very high dimensions 
and deal with new users or items appearing every second,
a matrix completion method that can be easily extended to unseen samples is of great significance.

In this work, we focus on the development of an algorithm 
that can recover the unknown entries of a partially observed matrix
even if some samples have not been seen during training.
Unlike previous studies~\cite{he17,hsieh17}, we \textit{only} employ explicit user feedback, 
that is, available matrix entries.
Our model relies on matrix factorization principles,
building upon a two-branch deep neural network architecture
to learn efficient latent representations of row and column samples.
We refer to our model as Neural Matrix Completion (NMC). 
Independent work \cite{xue17} has recently proposed a model similar to ours, which, however, 
focuses on predicting the personalized ranking over a set of items in recommender systems and employs \textit{both} explicit and implicit feedback. 
In contrast to \cite{xue17}, our model (\textit{i}) focuses on the matrix completion problem with emphasis on the extendability; and
(\textit{ii}) employs convolutional summarization layers, enabling its application to very high-dimensional matrices. 

The  rest of the paper is organized as follows:
Section \ref{sec:related} reviews the related work.
In Section \ref{sec:proposed}, we present the proposed neural network model,
show how the model can be extended to new samples, and discuss its application in high-dimensional matrices.
Section \ref{sec:experiments} includes experimental results in popular datasets.
Conclusions are drawn in Section \ref{sec:conclusions}.

\section{Related work}
\label{sec:related}
Neural networks have been proven effective in several domains
such as image classification~\cite{krizhevsky12}, sequence modeling \cite{sutskever14}, 
and inverse problems in signal processing \cite{nguyen17}.
In matrix completion, existing work involves autoencoders, graph convolutional networks and deep learning neural networks.
Autoencoder-based models~\cite{sedhain15, kuchaiev17} learn transformations 
from original row or column vectors to a latent space and decoders to predict the missing values. 
Geometric matrix completion models
employ graph convolutional networks to learn the feature vectors from column (or row) graphs~\cite{monti17} 
or bipartite user-item graphs~\cite{berg17}.
The CF-NADE (Collaborative Filtering - Neural Autoregressive Distribution Estimator) method \cite{zheng16}, on the other hand, 
learns directly the latent vectors from columns and rows.  
The Neural Collaborative Filtering (NCF)~\cite{he17} and Collaborative Metric Learning (CML)~\cite{hsieh17}  
utilize implicit feedback, i.e. interactions between users and items 
such as like, follows, shares, rather than explicit feedback, e.g. ratings.

A dominant idea behind many matrix completion algorithms 
is matrix factorization~\cite{Koren2009}.
Matrix factorization models are realizations of latent factor models.
They rely on the assumption that 
there exists an unknown representation of users and items in a low-dimensional  latent space
such that user-item interactions can be modelled as inner products in that space. 
Fitting a factor model to the data is a common approach in collaborative filtering~\cite{salakhutdinov07, Zhang2016}.
Recently,  neural networks have been employed to learn latent representations for matrix completion.
The work presented in~\cite{dziugaite15, he17} extends matrix factorization models 
by replacing the inner product with a non-linear function to model the interaction between row and column samples;
however, these models cannot be extended to samples unseen during training (see Section \ref{sec:generalizability} for further explanations).
\section{Extendable matrix completion}
\label{sec:proposed}
The proposed NMC model is a matrix factorization model.
The model learns vectors of the row and column samples using the available entries,
and predicts the missing entries using the inner product of the learned latent vectors.
The major challenge is, therefore,  to compute the mapping of users and items into latent vectors.
In our model, latent vector representations are obtained via a two-branch deep learning architecture.
Our model is trained with explicit feedback and no additional information is used.
Next, we present the model architecture that realizes this approach,
and discuss its ability to extend to new samples.

\subsection{Proposed Model}
\label{sec:model}
%
\begin{figure}[!t]
\centering
\includegraphics[width=0.7\linewidth]{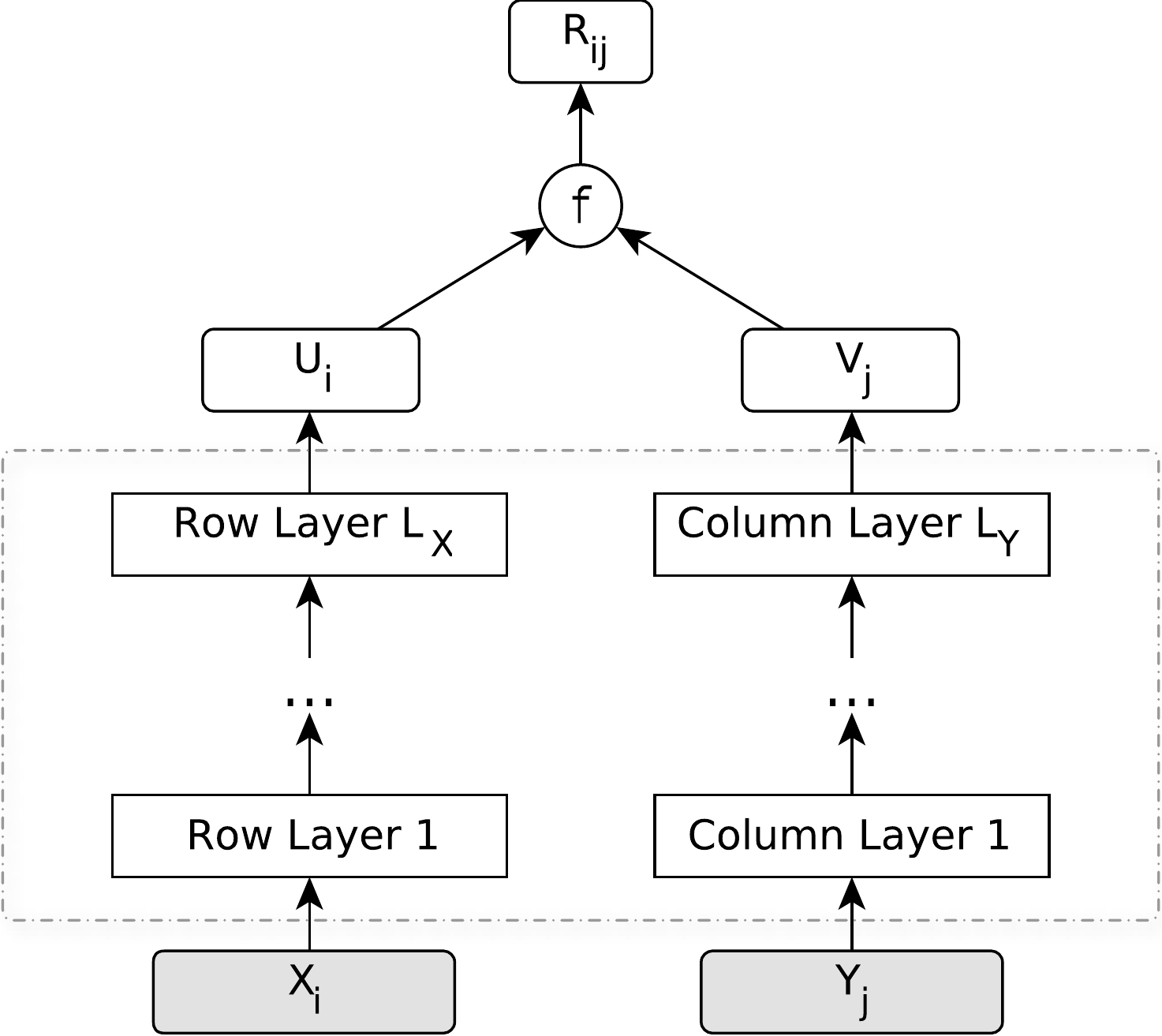}
\caption{The proposed two-stream neural network architecture for matrix completion.
$X_i$, $Y_j$ are input vectors, corresponding to the $i^\text{th}$ row and $j^\text{th}$ column of the original matrix.
The left and right branches of embedding layers consist of $L_X$ and $L_Y$ fully connected layers.
$U_i$, $V_j$ are the latent representations of $X_i$ and $Y_j$, respectively,
$f$ is a function to convert $\left( U_i,V_j \right)$ to the prediction $R_{ij}$.}
\label{fig:architecture}
\end{figure}
Consider a partially observed matrix $M \in \mathbb{R}^{n \times m}$,
and let $X_{i} \in \mathbb{R}^{m}, i = 1, \dots, n$, be the $i$-th row vector
and $Y_{j} \in \mathbb{R}^{n}, j = 1, \dots, m$, the $j$-th column vector. 
The proposed model for the prediction of the value  $R_{ij}$  at the $(i,j)$ matrix position
takes as input a pair of vectors $(X_{i}, Y_{j})$ and outputs
\begin{equation}
	R_{ij} = g\left( X_i, Y_j \right),
\label{modelbasic}
\end{equation}
where $R_{ij}$ is the predicted value at the $(i,j)$ position, 
and $g$ is a function reflecting the affinity between the $i$-th row and the  $j$-th column of the matrix.
A good model should follow the rule that similar row vectors and similar column vectors produce similar matrix values.

Following the matrix factorization strategy, 
we design a model that uses a mapping of row and column vectors into a latent $r$-dimensional factor space,
such that the obtained representations can provide predictions using the normalized inner product, i.e., the cosine similarity function.
Denoting by $U_i \in \mathbb{R}^{r}$, $V_j \in \mathbb{R}^{r}$ 
the latent representations of $X_i$ and $Y_j$, respectively,
the value  $R_{ij}$  at the $(i,j)$ matrix position is given by:
\begin{equation}
	R_{ij} = f\left( U_i, V_j \right)  = \dfrac{U_{i}^{T} V_{j}}{\left\|U_i \right\|_2 \left\|V_j \right\|_2} ~.
\end{equation}
Denoting the two embedding functions by $h_X$ and $h_Y$, 
the representations of $X_i$ and $Y_j$ in the latent space are  
$U_i = h_X(X_i)$ and $V_j = h_Y(Y_j)$.

In this work, we rely on the ability of deep neural networks to provide complex representations 
that can capture the relations between the underlying data.
The proposed architecture is presented in Fig.~\ref{fig:architecture}.
The two branches are designed to map row and column vectors into a shared latent space.
The embedding functions $h_X$ and $h_Y$ are realized by a number of $L_X$ and $L_Y$ fully connected layers, respectively,
each followed by a batch normalization layer \cite{ioffe15} and a ReLU activation function \cite{glorot11}.
To mitigate overfitting, we add a Dropout layer \cite{srivastava14} after all but the last hidden layers.
The use of batch normalization layers also helps control overfitting \cite{ioffe15}. 
We learn $h_X$ and $h_Y$ by fitting the observed data.
Our training set is created from the partially observed matrix $M$.
Specifically, we create two sets of samples, $\cal{X}$ and $\cal{Y}$, 
with $\cal{X}$ containing $n$ row samples $X_{i} \in \mathbb{R}^{m}$, $ i = 1, \dots, n$, 
and $\cal{Y}$ containing $m$ column samples $Y_{j} \in \mathbb{R}^{n}$, $j = 1, \dots, m$.
The inputs for our NMC model are taken from these two sample sets, $\cal{X}$ and $\cal{Y}$.

Since the cosine similarity between two vectors lies in $[-1,1]$, 
all entries $M_{ij} \in \left[ \alpha, \beta \right] $ in the original matrix $M$ 
are scaled into this range during training, according to
\begin{equation}
	M_{ij} = \dfrac{M_{ij} - \mu}{\mu - \alpha} ~, \quad \forall i,j,
	\label{eq:scaling}
\end{equation}
with $\mu = \left( \alpha + \beta \right) / 2$.
After the prediction, a re-scaling step is required to bring the estimated matrix $R$ to 
the same value range as $M$. 

We employ the mean square error (MSE) as the loss function to train NMC,
\begin{equation}
	L_\text{MSE} = \dfrac{1}{| \Omega_\text{tr} |}\sum_{ij \in \Omega_\text{tr}} (R_{ij} - M_{ij} )^{2},
	\label{eq:l2_loss}
\end{equation}
where $\Omega_\text{tr}$ is the set of indices of entries available during training and $| \Omega_\text{tr} |$ is its cardinality.

\subsection{Extending NMC to New Samples}
\label{sec:generalizability}
\begin{figure}[!t]
    \centering
    \includegraphics[scale=0.3]{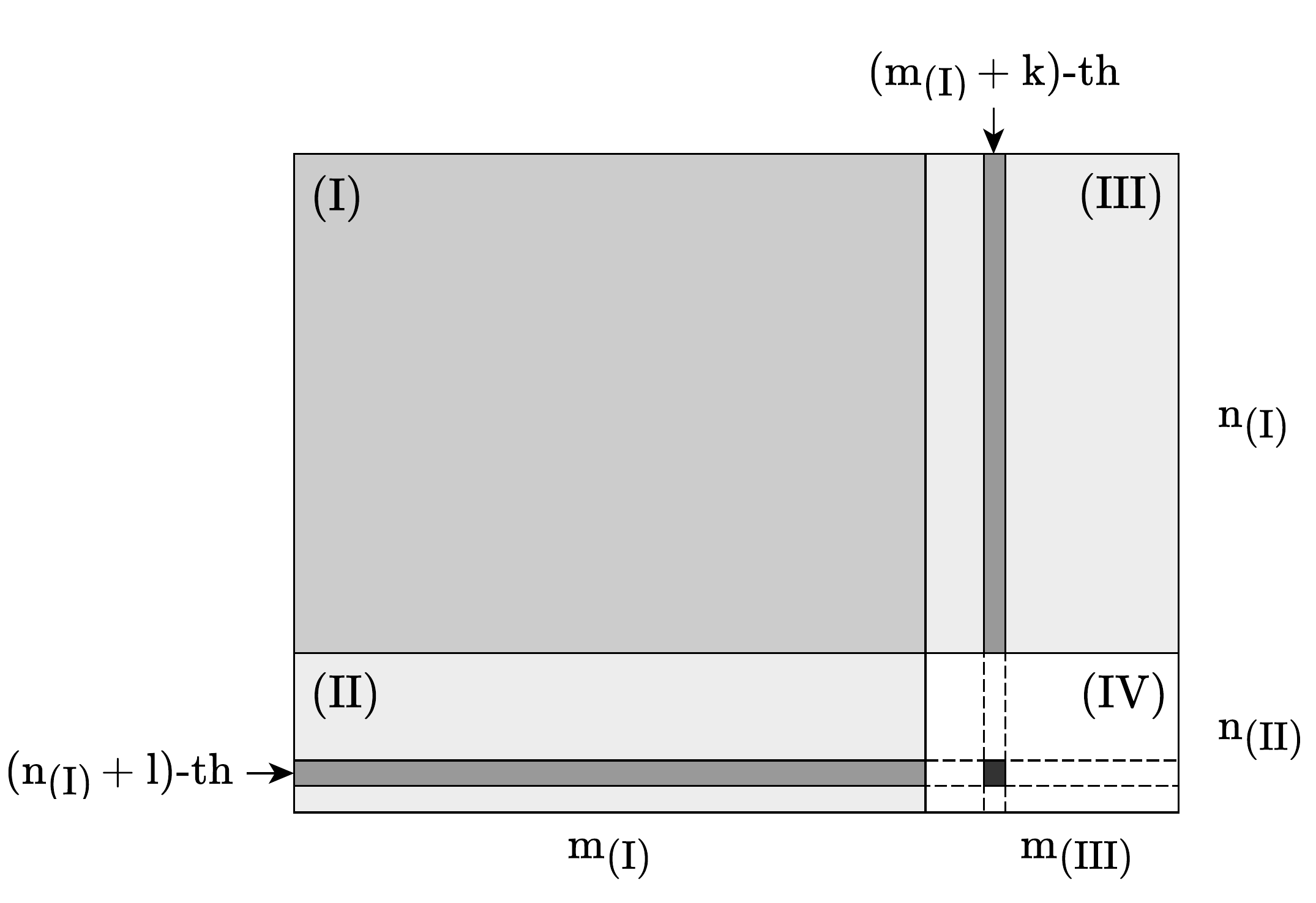}
    \caption{Extendability in matrix completion. 
    \textit{Dark shaded area} (I): rows and columns available during training.
    \textit{Light shaded areas} (II) and (III): entries corresponding to the interactions of 
    unseen rows and seen columns and vice versa. 
    \textit{White area} (IV): entries corresponding to the interactions of 
    unseen rows and unseen columns.}
\label{fig:rows_cols}
\end{figure}

A major problem of deep-neural-network-based 
models for matrix completion is related to their capability to be extended to 
samples unseen during training. 
An illustration of this problem is shown in Fig.~\ref{fig:rows_cols}.
The dark shaded area (I) corresponds to a submatrix $M_\text{(I)}$ of $M$ which is available during training. 
Only a small number of entries in $M_\text{(I)}$ are observed. 
Therefore, our model described in Section \ref{sec:model} can only be trained with input vectors 
$X_i \in \mathbb{R}^{m_{\text{(I)}}}$, $i=1,\dots,n_{\text{(I)}}$, 
and $Y_j \in \mathbb{R}^{n_{\text{(I)}}}$, $j=1,\dots,m_{\text{(I)}}$. 
The rows at the bottom and the columns to the right of $M$, which belong to light-shaded areas (II) and (III), 
are denoted as $M_\text{(II)}$ and $M_\text{(III)}$, respectively. 
They consist of new samples that are completely unseen during training. 
In recommender systems, these rows and columns represent new users and items.
Thus, the light-shaded areas (II) and (III) represent the interactions between new users (or items), with 
existing items (or users), while the white area (IV) represents the interactions of new users with new items. 
It should be noted that, even though $M_\text{(II)}$ , $M_\text{(III)}$ and $M_\text{(IV)}$ are completely unseen during training,
in this setting, they are not \textit{zero vectors}, but they contain partial observations.
Similar to $M_\text{(I)}$, the partial observations of $M_\text{(II)}$,  $M_\text{(III)}$  and $M_\text{(IV)}$ are used at testing,
where the task is to predict all the missing values from these observed ones. 

In existing deep-network-based methods, 
the models are trained and evaluated on the same area.
While this procedure enables measuring the accuracy of the predicted matrix, 
it does not measure how well a method can be extended to new rows and columns. 
%
NMC can provide predictions not only for unknown entries belonging to area (I) but also for entries in areas (II), (III) and (IV).
The most important feature of NMC compared to existing models
is that the functions transforming the original row and column vectors into the latent space are learned separately for rows and columns,
and the model architecture enables direct employment of the embedding functions for rows and columns unseen during training.

In recommender systems, predicting the unknown entries of a row belonging to area (II) 
is equivalent to providing recommendations for existing items to a user unseen during training.
In Fig.~\ref{fig:rows_cols}, the new user is represented by 
a new partially observed row vector, $X_{n_\text{(I)}+\ell}$, at the position $n_\text{(I)}+\ell$,  where $\ell = 1, \dots, n_\text{(II)}$.
The $(n_\text{(I)}+\ell)$-th user can interact with any column vector $Y_{j}$, $j=1,\dots, m_\text{(I)}$,
according to \eqref{modelbasic},
so as the model can fill in the missing entries at any position at the $(n_\text{(I)}+\ell)$-th row.
In a similar way, NMC can be extended to area (III).

Suppose now that we want to fill in an unknown entry at the $(n_\text{(I)}+\ell, m_\text{(I)}+k)$ position in area (IV), 
where $\ell = 1, \dots, n_\text{(II)}$, $k = 1, \dots, m_\text{(III)}$ 
(see Fig.~\ref{fig:rows_cols}).
Suppose that a row vector corresponding to the $(n_\text{(I)}+\ell)$-th new user is available and its dimension is $m_\text{(I)}+m_\text{(III)}$.
Suppose, also, that a column vector corresponding to the $(m_\text{(I)}+k)$-th new item is available and its dimension is $n_\text{(I)}+n_\text{(II)}$.
NMC ignores any observations in area (IV) and takes the first $m_\text{(I)}$ elements of $X_{n_\text{(I)}+\ell}$ row corresponding to ratings for the existing $m_\text{(I)}$ items,
and the first $n_\text{(I)}$ elements of $Y_{m_\text{(I)}+k}$ column corresponding to ratings of the existing $n_\text{(I)}$ users for the  $(m_\text{(I)}+k)$-th item
to form the input vectors; then, the model given by \eqref{modelbasic} can fill in the $(n_\text{(I)}+\ell, m_\text{(I)}+k)$ empty entry in area (IV).
\begin{table*}[t]
\centering
\caption{Matrix completion results on the ML-1M dataset~~\cite{harper15}.}
\label{table:exp_1_ml1m}
\begin{tabular}{ c | c | c | c | c | c | c | c | c }
\hline \hline
& \multicolumn{2}{|c|}{Area (I)} & \multicolumn{2}{|c|}{Area (II)} & \multicolumn{2}{|c|}{Area (III)} & \multicolumn{2}{|c}{Area (IV)} \\
\cline{2-9}
& RMSE & MAE & RMSE & MAE & RMSE & MAE & RMSE & MAE \\
\hline
U-CF-NADE-S \cite{zheng16} 			& $0.855$ & $0.671$ &    -    &    -    &    -    &    -    &    -    &    -    \\
\hline
I-CF-NADE-S \cite{zheng16} 			& $\textbf{0.839}$ & $\textbf{0.651}$ &    -    &    -    &    -    &    -    &    -    &    -    \\
\hline
U-Autorec \cite{sedhain15}			& $0.906$ & $0.722$ & $0.976$ & $0.781$ &    -    &    -    &    -    &    -    \\
\hline
I-Autorec \cite{sedhain15}			& $0.841$ & $0.662$ &    -    &    -    & $\textbf{0.856}$ & $\textbf{0.670}$ &    -    &    -    \\
\hline
Deep U-Autorec \cite{kuchaiev17}	& $0.889$ & $0.702$ & $0.969$ & $0.765$ &    -    &    -    &    -    &    -    \\
\hline
NMC-S						& $0.850$ & $0.675$ & $\textbf{0.883}$ & $\textbf{0.699}$ & $0.864$ & $0.685$ & $\textbf{0.904}$ & $\textbf{0.715}$ \\
\hline \hline
\end{tabular}
\end{table*}
\begin{table*}
\centering
\caption{Matrix completion results on the Netflix dataset~\cite{bennett07}.}
\label{table:exp_1_netflix}
\begin{tabular}{ c | c | c | c | c | c | c | c | c }
\hline \hline
& \multicolumn{2}{|c|}{Area (I)} & \multicolumn{2}{|c|}{Area (II)} & \multicolumn{2}{|c|}{Area (III)} & \multicolumn{2}{|c}{Area (IV)} \\
\cline{2-9}
& RMSE & MAE & RMSE & MAE & RMSE & MAE & RMSE & MAE \\
\hline
I-Autorec \cite{sedhain15}			& $\textbf{0.842}$ & $\textbf{0.655}$ &    -    &    -    & $\textbf{0.862}$ & $\textbf{0.671}$ &    -    &    -    \\
\hline
Deep U-Autorec \cite{kuchaiev17}	& $0.848$ & $0.662$ & $0.879$ & $0.689$ &    -    &    -    &    -    &    -    \\
\hline
NMC-S						& $0.856$ & $0.676$ & $\textbf{0.861}$ 	& $\textbf{0.680}$ & $0.873$ & $0.688$ & $\textbf{0.877}$ & $\textbf{0.692}$ \\
\hline \hline
\end{tabular}
\end{table*}
%

NMC employs a two-stream network architecture for matrix completion, similar to \cite{dziugaite15, hsieh17, he17}.
Nevertheless, the models \cite{dziugaite15, hsieh17, he17} are tied to the users and items available during training,
using one-hot vector representations for each user and each item
corresponding to the indices of rows and columns in the matrix $M_{\text{(I)}}$. 
In other words, the embedding function is a mapping from row and column indices to the latent representations. 
Hence, these methods cannot be extended to new rows and columns whose indices are not available during training. 
Comparing our method with Autorec proposed in \cite{sedhain15, kuchaiev17},
we incorporate both row and column vectors at the same time, while Autorec works with either rows or columns. 
This brings an advantage on the extendability of NMC compared to Autorec
(NMC can extend in both dimensions while Autorec only in one). 
Other recent neural-network-based model ~\cite{berg17}, 
even though achieves top performance in many benchmarks, cannot be extended to samples outside area (I). 
%
%
%
%
%
%
%
%
%
%
\subsection{Scaling NCM to High Dimensional Matrices}
Dimensionality is another major problem that has to be addressed by matrix completion methods;
in many settings the dimensions of the matrix of interest can be extremely large.
For example, the Netflix problem~\cite{bennett07} deals with matrix dimensions of the order of several thousands.
Directly applying the NMC model to extremely large matrices is not optimal,
due to the high dimensionality and sparsity of the inputs.

We propose the use of one or multiple summarization layers to reduce the
input dimensions before the embedding layers.
Each summarization layer is composed of a 1D convolutional layer, with a pre-defined number of filters
of adjustable kernel sizes, followed by
a batch normalization layer \cite{ioffe15} and a ReLU activation function \cite{glorot11}.
By properly configuring the number of filters and kernel size,
each summarization layer slides across the row and column vectors, and
summarizes them into denser vectors of lower dimensions.
We call this variant of NMC with summarization layers as NMC-S.
%
%
%
\section{Experimental results}
\label{sec:experiments}
%
We evaluate the proposed NMC-S model with experiments 
employing real matrices of varying dimensions and sparsity levels
and compare it with state-of-the-art methods.
Next, we present results involving the following two movie rating datasets:
one version of the MovieLens dataset \cite{harper15}
with one million available user ratings (ML-1M) and the Netflix dataset \cite{bennett07}. 

We run experiments on five random splits of each dataset. 
Each split involves partitioning the dataset into four parts, 
corresponding to areas (I) to (IV), as in Fig.~\ref{fig:rows_cols}.
We randomly shuffle the rows and columns of the given matrix, 
so that two splits are always different. 
Area (I) is assigned $80\%$ of the row and column samples. 
Following \cite{sedhain15, zheng16}, in each area
we randomly mark $90 \%$ of the available entries as observed; 
the remaining $10 \%$ are reserved for evaluation. 
The training set is formed only by the observed entries in area (I).
During evaluation, the model predicts the reserved test entries from the observed ones in all areas. 
The reserved test entries are used to calculate the prediction error.


For the evaluation of our model, we employ
the root mean square error (RMSE) and the mean absolute error (MAE)
defined as follows:
$\text{RMSE} = \sqrt{\sum_{ij \in \Omega_\text{eval}} (R_{ij} - M_{ij})^2 / \left| \Omega_\text{eval} \right|} $
$\text{MAE}  = \sum_{ij \in \Omega_\text{eval}} \left| R_{ij} - M_{ij} \right| / \left| \Omega_\text{eval} \right| $,
where $\Omega_\text{eval}$ is the set of indices corresponding to entries available for evaluation 
and $\left| \Omega_\text{eval} \right|$ represents the cardinality of $\Omega_\text{eval}$.

%

For the ML-1M dataset, in each branch of NMC-S model, we use two hidden layers
with $2048$ and $1024$ hidden units, respectively.
We employ a summarization layer of $32$ filters in both branches, 
with kernel size $32$ and $48$ and stride $16$ and $24$, respectively. 
It is worth mentioning that since the matrix dimension of the ML-1M dataset is not too large, 
the summarization layers employed do not necessarily reduce the dimensions of the input vectors. 
Since the number of training samples in this dataset is small, 
we employ a dropout regularizer with ratio $0.6$.
On the Netflix dataset, the matrix dimensions become very high. 
The NMC-S model is constructed with two embedding layers in each branch, 
both with hidden size of $2048$. The row branch has one summarization layer with filter size 
$128$ and stride $64$. The column branch has two summarization layers, 
with filter sizes $96$ and $64$ and strides $48$ and $32$, 
respectively. 

We compare the proposed NMC-S model against state-of-the-art matrix completion methods, namely, 
Autorec \cite{sedhain15}, Deep Autorec \cite{kuchaiev17} and CF-NADE \cite{zheng16}. 
%
We use the default parameters for the user-based autorec (U-Autorec), 
item-based autorec (I-Autorec) and CF-NADE models as in \cite{sedhain15,zheng16}. 
It should be noted that the results shown here are slightly different than those reported in \cite{zheng16, sedhain15}, 
since we train the models on a subset of the given matrices [area (I)].
Nevertheless, the relative performance ranking is consistent with \cite{zheng16, sedhain15}. 

Table \ref{table:exp_1_ml1m} presents the results for the ML-1M dataset.
In area (I), I-CF-NADE-S achieves the best performance, followed by I-Autorec and our NMC-S model.
NMC-S  outperforms U-Autorec and Deep U-Autorec in area (II). 
In area (III), I-Autorec has the best performance, yet followed closely by our NMC-S. 
It should be noted that, even though NMC-S has lower performance than I-CF-NADE-S and I-Autorec in 
area (I) and area (III), the performance difference is small while the overall accuracy gain in the other areas is significant. 
Furthermore, the proposed NMC-S model is the only one that can be extended to area (IV).

The results for the Netflix dataset are presented in Table \ref{table:exp_1_netflix}. 
On this dataset, we do not include the two CF-NADE models because of their high associated complexity. 
As can be seen, I-Autorec performs the best in areas (I) and (III), among the three models. 
The large number of training data improves the performance of Deep U-Autorec,  
which in this dataset achieves better results in both areas (I) and (II).
However, NMC-S is the only model that can be extended to all areas 
and delivers the best performance in areas (II) and (IV). 
It is worth mentioning that many techniques employed for Deep U-Autorec \cite{kuchaiev17}, 
such as dense re-feeding or heavy regularization, 
can also be used to boost the performance of NMC-S. 
Nevertheless, we leave this exploration for future work. 
%
%
%
%
\section{Conclusions}
\label{sec:conclusions}
We presented a novel matrix completion method, namely, NMC, 
which relies on the principles of matrix factorization.
Our model is realized by a two-branch neural network architecture
that maps row and column data into a joint latent space 
such that the relations between row and column samples 
can be modelled as inner products in that space.  
Our method can be extended to data unseen during training,
a feature that is of great significance in recommender systems
where new users or items appear every second. 
Easily applied to high-dimensional matrices, the proposed model
can be used to address well-known high-dimensional problems such as the Netflix problem.
Experiments  performed on real matrices of varying dimensions and sparsity levels have shown  
the effectiveness and robustness of our model 
with respect to the state of the art.

\bibliographystyle{IEEEtran}
\bibliography{mcicasspRefs}

\begin{thebibliography}{10}
\providecommand{\url}[1]{#1}
\csname url@samestyle\endcsname
\providecommand{\newblock}{\relax}
\providecommand{\bibinfo}[2]{#2}
\providecommand{\BIBentrySTDinterwordspacing}{\spaceskip=0pt\relax}
\providecommand{\BIBentryALTinterwordstretchfactor}{4}
\providecommand{\BIBentryALTinterwordspacing}{\spaceskip=\fontdimen2\font plus
\BIBentryALTinterwordstretchfactor\fontdimen3\font minus
  \fontdimen4\font\relax}
\providecommand{\BIBforeignlanguage}[2]{{%
\expandafter\ifx\csname l@#1\endcsname\relax
\typeout{** WARNING: IEEEtran.bst: No hyphenation pattern has been}%
\typeout{** loaded for the language `#1'. Using the pattern for}%
\typeout{** the default language instead.}%
\else
\language=\csname l@#1\endcsname
\fi
#2}}
\providecommand{\BIBdecl}{\relax}
\BIBdecl

\bibitem{Cao2014}
F.~Cao, M.~Cai, and Y.~Tan, ``Image interpolation via low-rank matrix
  completion and recovery,'' \emph{IEEE Transactions on Circuits and Systems
  for Video Technology}, vol.~25, no.~8, pp. 1261 -- 1270, 2015.

\bibitem{Ji2010}
H.~Ji, C.~Liu, Z.~Shen, and Y.~Xu, ``Robust video denoising using low rank
  matrix completion,'' \emph{IEEE Conference on Computer Vision and Pattern
  Recognition (CVPR)}, pp. 1791--1798, 2010.

\bibitem{Cheng2013}
J.~Cheng, Q.~Ye, H.~Jiang, D.~Wang, and C.~Wang, ``{STCDG}: An efficient data
  gathering algorithm based on matrix completion for wireless sensor
  networks,'' \emph{IEEE Transactions on Wireless Communications}, vol.~12,
  no.~2, pp. 850 -- 861, 2013.

\bibitem{yi12}
J.~Yi, T.~Yang, R.~Jin, A.~K. Jain, and M.~Mahdavi, ``Robust ensemble
  clustering by matrix completion,'' in \emph{IEEE International Conference on
  Data Mining (ICDM)}, 2012, pp. 1176--1181.

\bibitem{luo15}
Y.~Luo, T.~Liu, D.~Tao, and C.~Xu, ``Multiview matrix completion for multilabel
  image classification,'' \emph{IEEE Transactions on Image Processing},
  vol.~24, no.~8, pp. 2355--2368, 2015.

\bibitem{zheng16}
Y.~Zheng, B.~Tang, W.~Ding, and H.~Zhou, ``A neural autoregressive approach to
  collaborative filtering,'' in \emph{International Conference on Machine
  Learning (ICML)}, 2016, pp. 764--773.

\bibitem{monti17}
F.~Monti, M.~M. Bronstein, and X.~Bresson, ``Geometric matrix completion with
  recurrent multi-graph neural networks,'' \emph{arXiv preprint
  arXiv:1704.06803}, 2017.

\bibitem{dziugaite15}
G.~K. Dziugaite and D.~M. Roy, ``Neural network matrix factorization,''
  \emph{arXiv preprint arXiv:1511.06443}, 2015.

\bibitem{he17}
X.~He, L.~Liao, H.~Zhang, L.~Nie, X.~Hu, and T.-S. Chua, ``Neural collaborative
  filtering,'' in \emph{International Conference on World Wide Web (WWW)},
  2017, pp. 173--182.

\bibitem{candes09}
E.~J. Cand{\`e}s and B.~Recht, ``Exact matrix completion via convex
  optimization,'' \emph{Foundations of Computational Mathematics}, vol.~9,
  no.~6, p. 717, 2009.

\bibitem{Markovsky2012}
I.~Markovsky, \emph{Low Rank Approximation: {A}lgorithms, {I}mplementation,
  {A}pplications}.\hskip 1em plus 0.5em minus 0.4em\relax Springer, 2012.

\bibitem{Koren2009}
Y.~Koren, R.~Bell, and C.~Volinsky, ``{M}atrix {F}actorization {T}echniques for
  {R}ecommender {S}ystems,'' \emph{IEEE Computer}, vol.~42, no.~8, pp. 30--37,
  2009.

\bibitem{jin16}
C.~Jin, S.~M. Kakade, and P.~Netrapalli, ``Provable efficient online matrix
  completion via non-convex stochastic gradient descent,'' in \emph{Advances in
  Neural Information Processing Systems (NIPS)}, 2016, pp. 4520--4528.

\bibitem{chouvardas17}
S.~Chouvardas, M.~A. Abdullah, L.~Claude, and M.~Draief, ``Robust online matrix
  completion on graphs,'' in \emph{IEEE International Conference on Acoustics,
  Speech and Signal Processing (ICASSP)}, 2017, pp. 4019--4023.

\bibitem{hsieh17}
C.~K. Hsieh, L.~Yang, Y.~Cui, T.-Y. Lin, S.~Belongie, and D.~Estrin,
  ``Collaborative metric learning,'' in \emph{International Conference on World
  Wide Web (WWW)}, 2017, pp. 193--201.

\bibitem{xue17}
H.-J. Xue, X.~Dai, J.~Zhang, S.~Huang, and J.~Chen, ``Deep matrix factorization
  models for recommender systems,'' in \emph{International Joint Conference on
  Artificial Intelligence (IJICAI)}, 2017, pp. 3203--3209.

\bibitem{krizhevsky12}
A.~Krizhevsky, I.~Sutskever, and G.~E. Hinton, ``Imagenet classification with
  deep convolutional neural networks,'' in \emph{International Conference on
  Neural Information Processing Systems (NIPS)}, 2012, pp. 1097--1105.

\bibitem{sutskever14}
I.~Sutskever, O.~Vinyals, and Q.~V. Le, ``Sequence to sequence learning with
  neural networks,'' in \emph{Advances in Neural Information Processing Systems
  (NIPS)}, 2014, pp. 3104--3112.

\bibitem{nguyen17}
D.~M. Nguyen, E.~Tsiligianni, and N.~Deligiannis, ``Deep learning sparse
  ternary projections for compressed sensing of images,'' in \emph{IEEE Global
  Conference on Signal and Information Processing (GlobalSIP) [Available:
  arXiv:1708.08311]}, 2017.

\bibitem{sedhain15}
S.~Sedhain, A.~K. Menon, S.~Sanner, and L.~Xie, ``Autorec: Autoencoders meet
  collaborative filtering,'' in \emph{International Conference on World Wide
  Web (WWW)}, 2015, pp. 111--112.

\bibitem{kuchaiev17}
O.~Kuchaiev and B.~Ginsburg, ``Training deep autoencoders for collaborative
  filtering,'' \emph{arXiv preprint arXiv:1708.01715}, 2017.

\bibitem{berg17}
R.~v.~d. Berg, T.~N. Kipf, and M.~Welling, ``Graph convolutional matrix
  completion,'' \emph{arXiv preprint arXiv:1706.02263}, 2017.

\bibitem{salakhutdinov07}
R.~Salakhutdinov and A.~Mnih, ``Probabilistic matrix factorization,'' in
  \emph{International Conference on Neural Information Processing Systems
  (NIPS)}, 2007, pp. 1257--1264.

\bibitem{Zhang2016}
H.~Zhang, F.~Shen, W.~Liu, X.~He, H.~Luan, and T.-S. Chua, ``Discrete
  collaborative filtering,'' in \emph{International ACM SIGIR Conference on
  Research and Development in Information Retrieval}, 2016, pp. 325--334.

\bibitem{ioffe15}
S.~Ioffe and C.~Szegedy, ``Batch normalization: Accelerating deep network
  training by reducing internal covariate shift,'' in \emph{International
  Conference on Machine Learning (ICML)}, 2015, pp. 448--456.

\bibitem{glorot11}
X.~Glorot, A.~Bordes, and Y.~Bengio, ``Deep sparse rectifier neural networks,''
  in \emph{International Conference on Artificial Intelligence and Statistics
  (AISTATS)}, 2011, pp. 315--323.

\bibitem{srivastava14}
N.~Srivastava, G.~E. Hinton, A.~Krizhevsky, I.~Sutskever, and R.~Salakhutdinov,
  ``Dropout: a simple way to prevent neural networks from overfitting,''
  \emph{Journal of Machine Learning Research}, vol.~15, pp. 1929--1958, 2014.

\bibitem{harper15}
F.~M. Harper and J.~A. Konstan, ``The {M}ovielens datasets: History and
  context,'' \emph{ACM Transactions on Interactive Intelligent Systems},
  vol.~5, no.~4, pp. 19:1--19:19, Dec. 2015.

\bibitem{bennett07}
J.~Bennett and S.~Lanning, ``The {N}etflix prize,'' in \emph{KDD Cup and
  Workshop in conjunction with KDD}, 2007.

\end{thebibliography}

\end{document}